%% file: main.tex
\documentclass[runningheads]{llncs}
\usepackage[T1]{fontenc}
\usepackage{algorithm}
\usepackage{amsmath}
\usepackage{amssymb} 
\usepackage{graphicx}
\usepackage{subcaption}
\usepackage{algpseudocode}
\usepackage{booktabs}
\usepackage{multirow}
\usepackage{tabularx}

\begin{document}

\title{DDOT: A Derivative-directed Dual-decoder Ordinary Differential Equation Transformer for Dynamic System Modeling}
%
%
\author{
Yang Chang\inst{1}\orcidID{0009-0006-8877-8694}\thanks{Corresponding author} \and 
Kuang-Da Wang\inst{2}\orcidID{0009-0004-0846-8254} \and 
Ping-Chun Hsieh\inst{3}\orcidID{0000-0002-2072-8950} \and 
Cheng-Kuan Lin\inst{4}\orcidID{0000-0001-8530-967X} \and 
Wen-Chih Peng\inst{5}\orcidID{0000-0002-0172-7311}
}

\authorrunning{Yang Chang, Kuang-Da Wang et al.} 

\institute{National Yang Ming Chiao Tung University, Hsinchu, Taiwan \\
\email{\{nyms0390.cs10, gdwang.cs10\}@nycu.edu.tw}}

%
\maketitle              
\input{section/0-Abstract}

\input{section/1-Introduction}
\input{section/2-Related_Work}
\input{section/3-Methodology}

\input{section/4-Experiment}
\input{section/5-Conclusion}
%
%
%

%
%
%
\bibliographystyle{splncs04}
\bibliography{bibliography}

\end{document}

%% file: section/0-Abstract.tex
\begin{abstract}
Uncovering the underlying ordinary differential equations (ODEs) that govern dynamic systems is crucial for advancing our understanding of complex phenomena. Traditional symbolic regression methods often struggle to capture the temporal dynamics and intervariable correlations inherent in ODEs. ODEFormer, a state-of-the-art method for inferring multidimensional ODEs from single trajectories, has made notable progress. However, its focus on single-trajectory evaluation is highly sensitive to initial starting points, which may not fully reflect true performance. To address this, we propose the divergence difference metric (DIV-diff), which evaluates divergence over a grid of points within the target region, offering a comprehensive and stable analysis of the variable space. Alongside, we introduce DDOT (Derivative-Directed Dual-Decoder Ordinary Differential Equation Transformer), a transformer-based model designed to reconstruct multidimensional ODEs in symbolic form. By incorporating an auxiliary task predicting the ODE’s derivative, DDOT effectively captures both structure and dynamic behavior. Experiments on ODEBench show DDOT outperforms existing symbolic regression methods, achieving an absolute improvement of 4.58\% and 1.62\% in $P_{R^2 > 0.9}$ for reconstruction and generalization tasks, respectively, and an absolute reduction of 3.55\% in DIV-diff. Furthermore, DDOT demonstrates real-world applicability on an anesthesia dataset, highlighting its practical impact.\footnote{The codebase will be released in the camera-ready version.}

\keywords{Symbolic Regression \and Ordinary Differential Equation.}
\end{abstract}

%% file: section/1-Introduction.tex
\section{Introduction}
In recent years, machine learning (ML) has significantly transformed scientific research methodologies. 
Dynamic systems modeling, that is, model systems evolve over time according to underlying physical, biological, or chemical laws, also benefit from the advancement of ML. 
Accurately capturing the temporal dynamics and interactions between variables in such systems is crucial to understanding and predicting their behavior.
Approaches such as neural ordinary differential equations (NODEs) \cite{NODE} leverage ML to model dynamic systems, but often result in black-box models that lack transparency and interpretability. This consideration leads to symbolic regression (SR), which generates interpretable mathematical expressions.
Traditional SR primarily employs genetic programming (GP) to derive predictive functions from observed data \cite{BackToTheFormula,koza1994genetic,rudin2019stop,schmidt2009distilling}. Although GP has been widely used to construct symbolic trees that link input variables $X$ to output variables $Y$, it focuses predominantly on identifying single static equations. 
Consequently, it does not capture the temporal dynamics and intervariable correlations inherent in ODEs. 
Moreover, GP suffers from significantly long inference times that grow exponentially with the length of the expression, since it cannot efficiently exploit observed data as it requires rerunning the entire process for each new dataset. In contrast, ML-based methods \cite{DeepSymbolicRegression} train a model once on a large dataset, enabling much quicker inference times for new data.

ODEFormer \cite{ODEFormer} was introduced as the first transformer-based SR method for inferring ODEs. It evaluates performance through \textbf{reconstruction} and \textbf{generalization} tasks. The reconstruction task compares the ground truth trajectory with the trajectory obtained by integrating the predicted ODE from the same initial condition over the same interval as the ground truth. The generalization task involves integrating both the ground truth and the inferred ODE for a new, different initial condition over the same interval and comparing the resulting trajectories. However, we argue that these tasks only partially reflect the performance of the predicted ODE, as they rely on single trajectories, which are highly sensitive to the choice of the initial point. 

\begin{figure*}[t]
\centering
\includegraphics[width=0.95\textwidth]{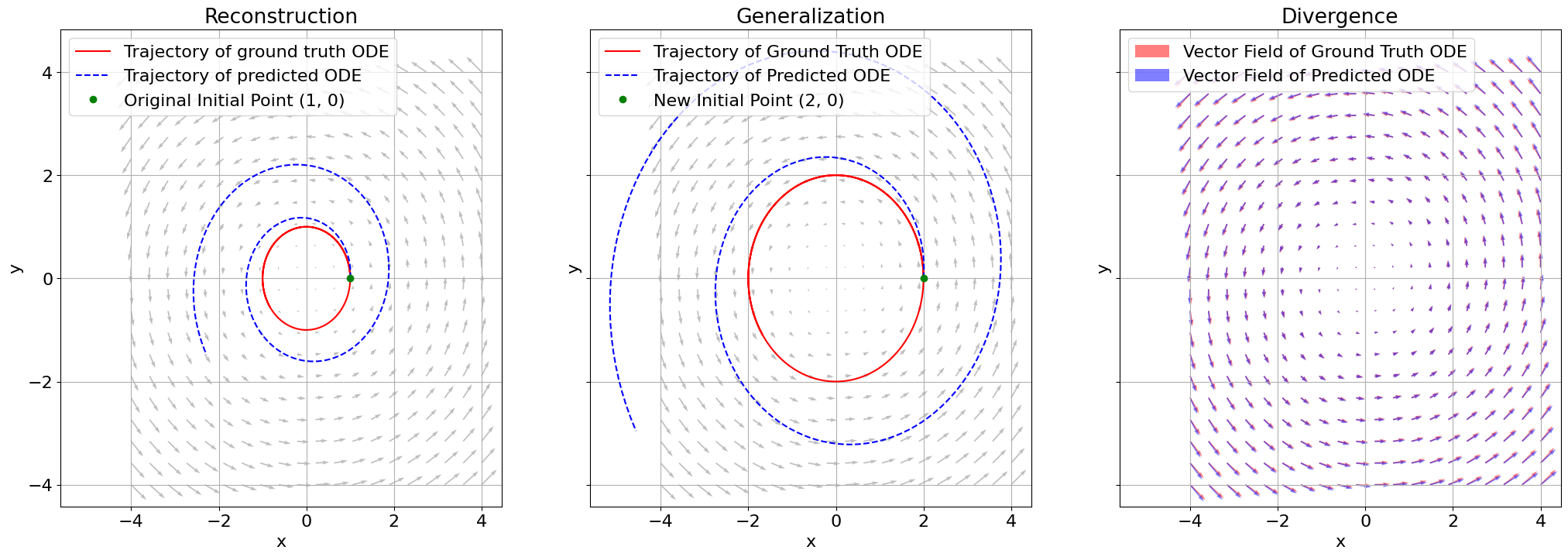}
\caption{\textbf{Evaluation Task Comparison}: Divergence offers a more thorough comparison across the entire variable space.}
\label{fig:intro_exp}
\end{figure*}

To address these limitations, we introduce a new metric: divergence difference (DIV-diff). In vector calculus, divergence is a vector operator that acts on a vector field, producing a scalar field that quantifies the field's source at each point. This metric is calculated by integrating the ODE over a grid of points within the targeted area and applying the resulting vector field to the ODE. 
Unlike single-trajectory analysis, DIV-diff is independent of initial value selection and provides a broader, targeted evaluation of the variable space.
Figure \ref{fig:intro_exp} illustrates the differences between reconstruction, generalization tasks, and divergence calculation. Consider the following ODEs: 

\begin{equation}
    \text{Ground Truth ODE: }
    \begin{cases} 
        \frac{dx}{dt} = -y \\ 
        \frac{dy}{dt} = x
    \end{cases}, \quad 
    \text{Predicted ODE: }
    \begin{cases} 
        \frac{dx}{dt} = -y + 0.1x \\ 
        \frac{dy}{dt} = x + 0.1y
    \end{cases}
\end{equation}

For the reconstruction and generalization tasks, the red line represents the roll-out trajectory of the ground truth ODE integrated from initial points $(1, 0)$ and $(2, 0)$, respectively, while the blue line represents the roll-out sequence of the predicted ODE. 
Trajectories from the same starting point can diverge significantly over time due to accumulated errors, and different starting points yield varying roll-out trajectories. This highlights the need to evaluate performance across the entire variable space rather than relying on a single trajectory.
In contrast, divergence calculates the scalar of the vector field at each grid point, offering a more comprehensive analysis of the variable space without compounding of errors. This makes the DIV-diff metric more reliable.
In this study, we further propose DDOT, a derivative-directed dual-decoder transformer model designed to infer ODEs from temporal data. 
DDOT incorporates an auxiliary task specifically aimed at improving performance on the divergence difference metric. 
Besides predicting the symbolic form of the ODE, DDOT also estimates its derivative. Directly optimizing the divergence is computationally complex, so we approximate it by predicting the roll-out trajectory of the ODE. Ideally, repeated predictions of the trajectory would converge to the optimal solution, effectively mirroring the original task. This approach is more efficient and still significantly improves overall model performance.
In summary, our contributions are as follows:
\begin{itemize}
\item We identify limitations in current evaluation methods and introduce divergence difference as a new metric for more comprehensive assessment.
\item We propose DDOT, a novel model that infers ODEs from temporal data and achieves superior performance across all evaluated baselines.
\item We demonstrate the potential of DDOT in real-world applications, showing that the inferred ODEs offer both predictive accuracy and interpretability.
\end{itemize}

%% file: section/2-Related_Work.tex
\begin{figure*}[h]
\centering
\includegraphics[width=0.95\textwidth]{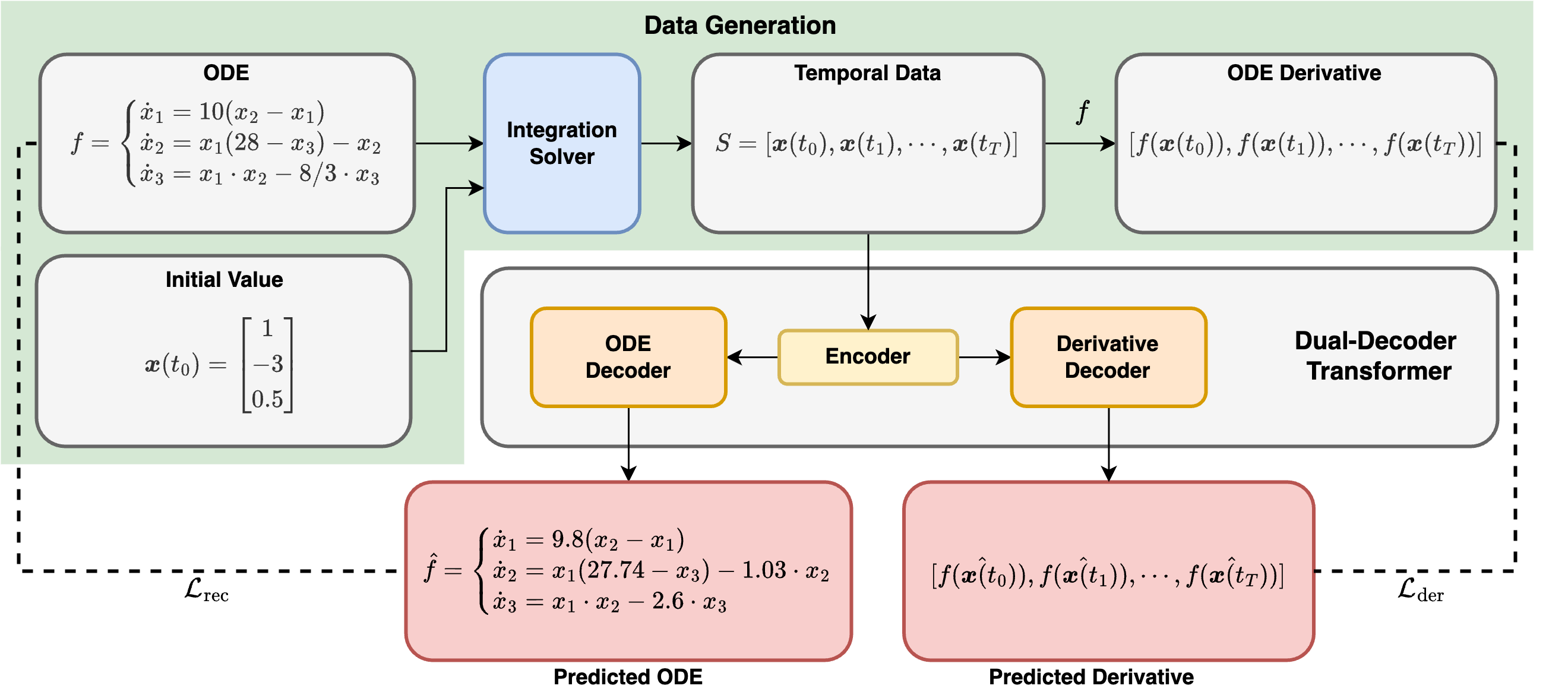}
\caption{\textbf{Sketch of the training structure of DDOT.} Illustrated with an example ODE: Lorenz System.}
\label{fig:model_structure}
\end{figure*}

\section{Related Work}
\subsection{Modeling Temporal Data}
Time Series Forecasting shares a similar goal with dynamic system modeling. However, while dynamic system modeling often aims to understand and simulate the underlying mechanisms and interactions within a system, time-series forecasting primarily focuses on analyzing historical data patterns to predict future values. Methods like DeepAR \cite{DeepAR} and Neural ODE (NODE) \cite{NODE} demonstrate the power of neural networks in this field, but their black-box nature presents significant challenges for interpretability.

\subsection{Symbolic regression} 
Symbolic regression (SR) is a powerful method for achieving interpretability by identifying a mathematical function that accurately models the data. Traditionally, SR is used to infer a static function $g$ from observations $(x, g(x))$, a process referred to as \textit{functional SR} by \cite{ODEFormer}. However, when modeling dynamic systems, the objective shifts to characterizing the time derivative $(\dot{x}=f(x))$ from trajectory data $(t, f(t))$, known as \textit{dynamic SR}. In this context, $[f(t)]$ may exhibit noise, which adds complexity to the modeling process. SR faces challenges due to its sensitivity to inaccuracies in labels, primarily because of the vast search space it must explore \cite{DBLP:conf/ppsn/AgapitosBO12}. In recent years, deep learning has been leveraged to enhance symbolic regression (SR) approaches. In AI Feynman \cite{AI-feynman}, neural networks identify simplifying properties in datasets, allowing the recursive definition of subproblems for SR. \cite{E2ESR} pointed out that the two-step procedure commonly used in SR (skeleton prediction followed by fitting constants) could suffer from an insufficient supervision problem and proposed a transformer-based end-to-end SR model. 
\cite{DeepSymbolicRegression} employed a recurrent neural network to emit a distribution over tractable mathematical expressions and trained the network with a novel risk-seeking policy gradient to avoid the backpropagation problem. 

\subsection{Dynamic Symbolic Regression}
However, the aforementioned examples focus mainly on functional SR. If we adapt these methods to compute $f'(t)$ using numerical solvers, they may introduce computational errors and precision issues, potentially leading to suboptimal performance. To address these challenges, \cite{ODEFormer} introduced ODEFormer, which utilizes a transformer-based model to predict ODE functions. 
While transformers are renowned for their ability to handle sequence-to-sequence tasks, their attention mechanism operates at the token level, which may not fully capture the dynamics of ODEs in variable space. 
Furthermore, \cite{ODEFormer} evaluated performance based on a single trajectory roll-out, which is highly sensitive to initial value selection, leading to significant variations in trajectories and making it challenging to accurately assess the predicted ODE's true performance.
In this study, we propose a stable evaluation metric, Divergence Difference (DIV-diff), which is independent of initial value selection. Additionally, we introduce an auxiliary task that aims to predict $f'(t)$ as a supplementary objective to further improve performance.

%% file: section/3-Methodology.tex
\section{Problem Formulation}
Denote the observations as $\boldsymbol{S} = {\boldsymbol{x}(t)}_{t\in T}$, where each $\boldsymbol{x}(t) \in \mathbb{R}^d$ signifies the state of the system at time $t$ with dimension $d$. Our aim is to develop a model capable of accurately predicting the mathematical formulation of the governing ODE $\frac{d\boldsymbol{x}}{dt} = f$. In addition, the model will address an auxiliary objective of predicting the derivative sequence $[f(\boldsymbol{x}(t))]$.

\section{Methodology}
The primary goal of this research was to derive symbolic representations of ordinary differential equation (ODE) systems using observed temporal data.
The DDOT training architecture, shown in Figure \ref{fig:model_structure}, includes an ODE data generation module and a Dual-Decoder Transformer model. The model takes temporal data (i.e., trajectories) generated by ODE integration as its input, and outputs both the predicted ODE and its derivative.
In the following sections, we describe the data generation process of our training dataset followed by introducing the architecture of the DDOT and its training method. 

\subsection{Data Generation}
We train our model using a large dataset of generated ODEs. Following the approach outlined in \cite{ODEFormer,E2ESR}, the generation process includes randomly generating ODEs and subsequently integrating them to produce sequential data. These sequential data are then fed into the generated ODEs to calculate $f(\boldsymbol{x}(t_i))$, the derivative of the state at each time point $t_i$.

\noindent\textbf{Generating ODEs.}
In prior approaches \cite{DLforSymbolicMath}, mathematical expressions are represented as trees, with operators as internal nodes and operands as leaves, ensuring a one-to-one mapping between trees and sequences. While we omit the detailed generation process, diversity is ensured through the random generation of constants, making each function—and by extension, the entire ODE system—distinctly sampled.

\noindent\textbf{Integrating ODEs.}
Once an ODE is sampled, we perform numerical integration to generate sequential data. Starting with the random generation of an initial condition $\boldsymbol{x}_0$, we integrate the ODE using the numerical solver \texttt{scipy.integrate\allowbreak.solve\_ivp} provided by SciPy \cite{SciPy} to obtain sequence $[\boldsymbol{x}(t_i)]$, where $\boldsymbol{x}(t_i)$ is the state of the system at time $t_i$. Given the random nature of ODE generation, there is the possibility of improper integration. If integration throws errors or exceeds one second, the sample is discarded.

\noindent\textbf{Computing Derivative.}
After integration, we plug the obtained sequence $[\boldsymbol{x}(t_i)]$ back into the ODE to evaluate the right-hand side of the ODE at each time point $t_i$:
\begin{equation}
\frac{dx}{dt} \bigg|_{t_i} = f(t_i, \boldsymbol{x}(t_i)).
\end{equation}
Here, $f$ is the function defining the ODE. Physically, we obtain the instantaneous rate of change of the system's state at each point in time, as predicted by the generated ODE.

\noindent\textbf{Tokenizing ODEs.}
For tokenizing mathematical expressions, we closely follow the methodology established in previous work \cite{ODEFormer}. Expressions are converted into prefix notation to eliminate ambiguity. The vocabulary includes tokens for operators, variables, special symbols (such as ``$|$", used to separate equations), and components for tokenizing floating-point numbers. Following the approach in \cite{LinearAlgebraWithTransformers}, floating-point numbers are broken down into three components: sign, mantissa, and exponent. For example, the float $3.14$ is tokenized as three tokens: $[+, \text{N0314}, \text{E-2}]$. The ODE:
\begin{equation}
    \begin{cases}
        \dot{x_0}=-x_1+0.1*x_0, \\
        \dot{x_1}=x_0+0.1*x_1.
    \end{cases}
\end{equation}
is tokenized as 
\begin{equation}
  \begin{split}
[ \; &\text{add}, -, x_1, \text{mul}, +, \text{N0001}, \text{E-1}, x_0, |, \\
 &\text{add}, x_0, \text{mul}, +, \text{N0001}, \text{E-1}, x_1 \; ].
  \end{split}
\end{equation}

\subsection{Model Architecture}
The proposed DDOT (Derivative-directed Dual-decoder Ordinary Differential Equation Transformer) model is based on the sequence-to-sequence Transformer framework \cite{AttentionIsAllYouNeed}. Figure \ref{fig:model_structure} illustrates our proposed training method. Specifically, DDOT takes ODE rolled-out trajectories as input and uses two decoders: an ODE decoder and a derivative decoder, to predict both the underlying ODE and its derivative. Compared to ODEFormer \cite{ODEFormer}, DDOT's additional decoder, predicting the derivative sequence, enhances performance by capturing system dynamics.

\noindent\textbf{Dual-decoder Structure.}
In DDOT, the \textit{ODE decoder} predicts the symbolic representation of the ODE, while the \textit{derivative decoder} predicts the sequence $[f(\boldsymbol{x}(t_i))]$, capturing both the symbolic form and dynamic behavior in the state space.
While SR models aim to identify the ODE accurately, they often struggle to capture the system's dynamic evolution. By incorporating a derivative decoder, our model learns both the equation and its behavior, offering a more holistic understanding of the system.

\noindent\textbf{Loss function.} 
Each decoder generates a sequence, and the loss for both the predicted ODE ($\mathcal{L}_\text{rec}$) and its derivative ($\mathcal{L}_\text{der}$) is computed using cross-entropy loss:
\begin{equation}
    \mathcal{L}=-\frac{1}{N}\sum_{j=1}^N\sum_i y_{j,i}\log\hat{y}_{j,i}
\end{equation}
The total model loss is the sum of $\mathcal{L}_\text{rec}$ and $\mathcal{L}_\text{der}$.
Since directly learning divergence is impractical, we approximate it by predicting derivatives and apply a cross-entropy loss, termed divergence difference loss ($\mathcal{L}_\text{der}$), for training. By estimating derivatives at each trajectory point, this approach approximates the true ODE vector field, resulting in a ground-truth divergence closely aligning with the true divergence. This combined loss ensures that the model captures both the ODE's structure and its dynamic behavior across the variable space.

%% file: section/4-Experiment.tex
\section{Experiments}

\subsection{Experimental Setup}
\noindent\textbf{Benchmarks.}
We use ODEBench, introduced by \cite{ODEFormer}, as our benchmark. ODEBench significantly extends the Strogatz dataset from the Penn Machine Learning Benchmark (PMLB) \cite{PMLB}, which includes only seven two-dimensional ODEs, each integrated with four different initial conditions. While the Strogatz dataset is limited by its focus on 2D systems, ODEBench includes 63 ODEs across various dimensions (1D: 23, 2D: 28, 3D: 10, 4D: 2), incorporating well-known real-world systems. This expanded dataset provides a more rigorous and diverse evaluation framework, enabling a comprehensive assessment of model performance across a broader range of dynamic systems. Therefore, ODEBench is used as the primary dataset in our experiments.

\noindent\textbf{Implementation Details.}
Our model is configured with 16 attention heads and a 512-dimensional embedding space. The architecture is asymmetric, consisting of 4 layers in the encoder and 12 layers in the decoder. We optimize the model using cross-entropy loss and the Adam optimizer, with the learning rate gradually increasing from $10^{-7}$ to $2 \times 10^{-4}$ over the first 10,000 steps, followed by a cosine decay over the next 300,000 steps. The entire training process requires approximately 600,000 optimization steps and takes around 6 days to complete.

\noindent\textbf{Baseline Methods.}
We evaluate DDOT against state-of-the-art dynamic symbolic regression methods, including ODEFormer, SINDy, and ProGED, focusing on their suitability for modeling time-dependent systems. ODEFormer leverages transformers for sequence-to-sequence translation, SINDy enhances interpretability through sparsity in regression-based models, and ProGED utilizes probabilistic grammar with Monte Carlo sampling. DDOT advances transformer-based approaches by integrating sequence-to-sequence translation with derivative prediction.

\noindent\textbf{Evaluation Metrics.}
For the comparison metrics, we include the following: 1) $P_{R^2>.9}$: Following ODEFormer\cite{ODEFormer}, we introduce the probability of $R^2$ above 0.9 as a metric. This measures the likelihood that the model achieves a high level of explanatory power. 2) DIV-diff: This proposed metric evaluates the \textit{difference in divergence} between the predicted and actual vector fields. To quantify this, we calculate the Root Logarithm Mean Squared Error (RLMSE) of the divergence difference, as there is no established method for this measurement.
It is important to note that $P_{R^2>.9}$ evaluate trajectory alignment, making them suitable only for reconstruction and generalization tasks. To assess the system's dynamic behavior more comprehensively, DIV-diff evaluates the entire vector field instead of a single trajectory, offering a deeper insight into the model's performance.


\begin{table}
\small
\caption{Quantitative results.}
\label{tab:quant_result}
\centering
\scriptsize
\resizebox{\columnwidth}{!}{%
\begin{tabular}{@{}l|llllll|llllll@{}}
\toprule
\multirow{2}{*}{\textbf{Method}} & \multicolumn{6}{c|}{\bf Reconstruction}  & \multicolumn{6}{c}{\bf Generalization} \\ 
\cmidrule(l){2-13} 
                & noise=0        & 0.01           & 0.02           & 0.03           & 0.04           & 0.05           & noise=0        & 0.01           & 0.02           & 0.03           & 0.04           & 0.05           \\ \midrule
ProGED          & 0.403          & 0.355          & 0.355          & 0.371          & 0.306          & 0.339          & 0.145          & 0.113          & 0.129          & 0.097          & 0.145          & 0.113          \\
SINDy           & 0.000          & 0.016          & 0.000          & 0.000          & 0.000          & 0.000          & 0.065          & 0.000          & 0.000          & 0.000          & 0.000          & 0.000          \\
ODEFormer       & \underline{0.581} & \underline{0.581} & \underline{0.548} & \underline{0.581}  & \underline{0.532} & \underline{0.516} & \textbf{0.210}   & \underline{0.226} & \textbf{0.226} & \underline{0.177} & \underline{0.210} & \underline{0.194} \\
DDOT            & \textbf{0.613} & \textbf{0.629} & \textbf{0.613} & \textbf{0.613} & \textbf{0.581} & \textbf{0.565} & \underline{0.194} & \textbf{0.242} & \textbf{0.226} & \textbf{0.210} & \textbf{0.226} & \textbf{0.242} \\ \bottomrule
\end{tabular}
}
\end{table}

\subsection{Quantitative Results.}
Experimental results for reconstruction and generalization tasks are shown in Tables \ref{tab:quant_result}. We include $P_{R^2>.9}$, which measures the explanatory power of the model.
Table \ref{tab:Divergence} presents the performance of the DIV-diff in predicting the divergence between the predicted and ground truth ODEs. Performance was evaluated across five different noise levels to assess the impact of varying degrees of noise on model accuracy.

\noindent\textbf{Reconstruction Task.}
For the reconstruction task, the model reconstructs the original trajectory from the dataset. DDOT consistently outperforms all baselines across noise levels, with average absolute improvements of $4.58$\% in $P_{R^2 >.9}$ over the second-best ODEFormer. These results demonstrate DDOT's superior reconstruction capability and the effectiveness of the derivative decoder in learning dynamic behavior in the state space. Compared to ProGED and SINDy, both DDOT and ODEFormer exhibit greater robustness to noise, maintaining high performance even at the highest noise levels.

\noindent\textbf{Generalization Task.}
In the generalization task, a new initial point is randomly selected, and both the ground truth and predicted ODEs are integrated to assess the similarity of the resulting trajectories. This task is more challenging than reconstruction and is highly sensitive to the choice of initial values.
We highlight the significant drop in performance across all models due to the task's inherent sensitivity. Although DDOT consistently ranks first or second in $P_{R^2 >.9}$, these results emphasize the need to evaluate performance across the entire variable space rather than a single trajectory. Therefore, we propose using DIV-diff to mitigate sensitivity to the choice of initial values.

\begin{table}
\caption{Divergence difference from ground truth between baselines and DDOT.}
\label{tab:Divergence}
\centering
\begin{tabular}{@{}l|llllll@{}}
\toprule
\textbf{Method} & noise=0        & 0.01           & 0.02           & 0.03           & 0.04           & 0.05           \\ \midrule
ProGED          & 3.271           & 4.772           & 3.671           & 4.421           & 4.274            & 4.517            \\
SINDy           & 5.486          & 7.613          & 7.946          & 8.045          & 8.022          & 8.040          \\
ODEFormer       & \underline{2.870} & \textbf{2.703} & \underline{2.779} & \underline{2.881} & \underline{2.944}   & \underline{2.819} \\
DDOT            & \textbf{2.799} & \underline{2.973}  & \textbf{2.710} & \textbf{2.681} & \textbf{2.840} & \textbf{2.780} \\ \bottomrule
\end{tabular}
\end{table}

\noindent\textbf{Divergence Difference.}
The DIV-diff metric measures the difference in divergence between two ODEs, evaluated over a selected variable space, providing a more comprehensive measure of overall similarity. In Table \ref{tab:Divergence}, DDOT outperforms other models across most noise levels on this metric, demonstrating its ability to capture the true behavior of dynamic systems.

\subsection{Divergence Visualization.}
\begin{figure}
    \centering
    \begin{subfigure}[c]{0.70\textwidth}
        \centering
        \includegraphics[width=\textwidth]{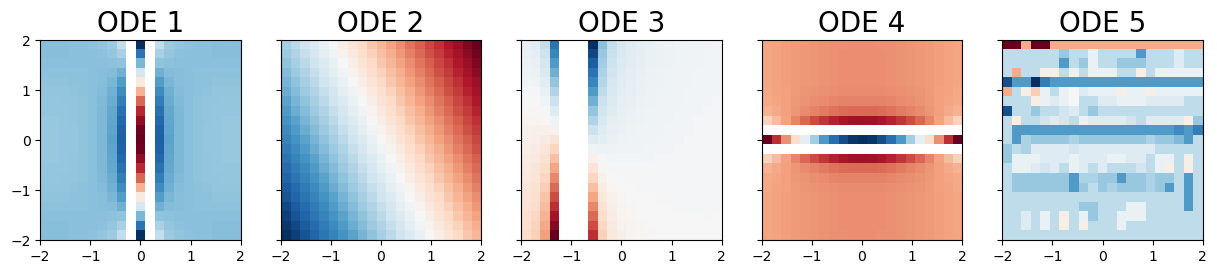}
        \caption{Ground Truth}
        \label{fig:sub1}
    \end{subfigure}
    \hspace{5cm}
    \begin{subfigure}[c]{0.70\textwidth}
        \centering
        \includegraphics[width=\textwidth]{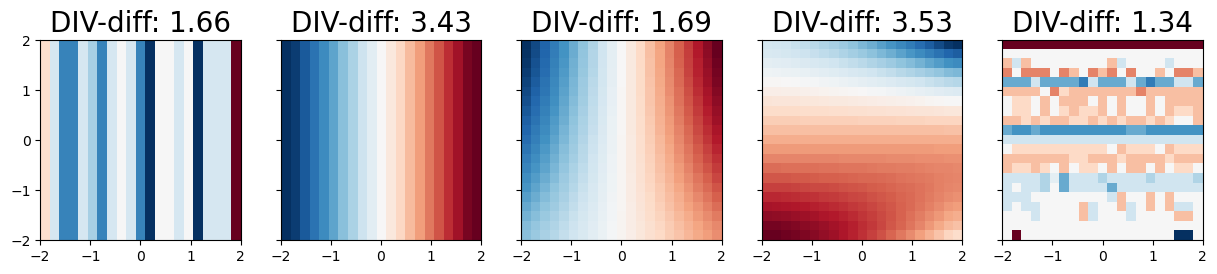}
        \caption{Predicted by DDOT}
        \label{fig:sub2}
    \end{subfigure}
    \hspace{5cm}
    \begin{subfigure}[c]{0.70\textwidth}
        \centering
        \includegraphics[width=\textwidth]{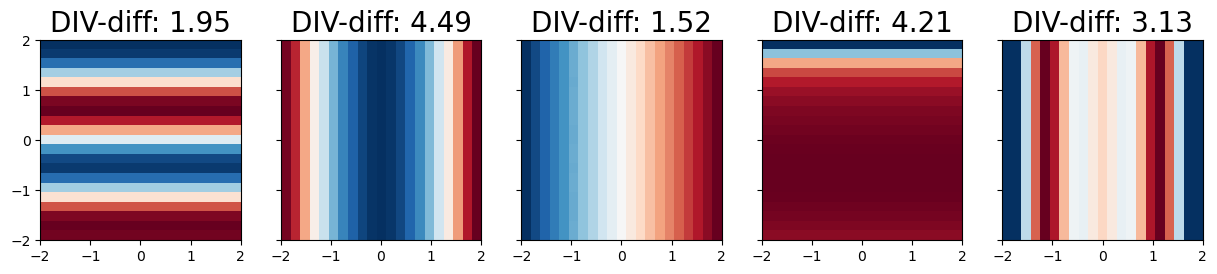}
        \caption{Predicted by ODEFormer}
        \label{fig:sub3}
    \end{subfigure}
    \caption{\textbf{Divergence Visualization} for 5 ODEs from the ODEBench dataset shows that red indicates lower divergence and blue higher. The color patterns reveal that DDOT captures the true ODE dynamics more effectively than ODEFormer, with lower DIV-diff values indicating better performance.}
    \vspace{-7pt}
    \label{fig:vis_div}
\end{figure}

To illustrate the performance on the DIV-diff metric, we visualized the divergence of 5 ODEs from the ODEBench dataset in Figure \ref{fig:vis_div}, comparing the ground truth ODEs with those predicted by DDOT and ODEFormer. The first row shows the divergence of the ground truth ODEs, the second row displays the divergence of ODEs predicted by DDOT, and the third row presents those predicted by ODEFormer. The difference in divergence for each ODE is indicated above each figure. A more similar divergence graph corresponds to a lower div value, reinforcing the reliability of this metric through visual comparison. The visual comparison also reveals that DDOT better captures the ground truth ODE divergence patterns compared to ODEFormer.

\subsection{Case Studies}
\noindent\textbf{Inference ODE from ODEBench Dataset.}
To provide a deeper understanding of DDOT's capability in inferring underlying dynamic systems, we compare the predicted ODEs with the ground truth ODEs from ODEBench. Table \ref{tab:case_study_odebench} highlights examples of predicted equations alongside their respective ground truth equations for selected benchmark systems. We demonstrate that the predicted ODEs closely match the ground truth ODEs for lower-dimensional systems, with parameters estimated to be very close to the actual values. However, as the dimensionality of the system increases, prediction inaccuracies also grow. For the chaotic Lorenz system, only part of the predicted equations is structurally similar to the ground truth, highlighting the challenges of modeling chaotic systems.

\noindent\textbf{Inference ODE from the Anesthesia Dataset.}
We showcase DDOT's application in anesthesia management by modeling interactions between anesthetic dosage and physiological indicators, including Propofol, heart rate, SBP, MBP, DBP, and BIS. The model dynamically predicts the BIS index, a key anesthesia metric, by constructing and integrating ODEs based on patient-specific initial states. Table \ref{tab:anes_vars} summarizes key variables related to anesthesia surgery, along with their descriptions and inferred ODE functions, highlighting the potential for uncovering relationships between these variables. This case study highlights DDOT’s ability to derive interpretable symbolic functions from real-world data, providing actionable insights for clinical decision-making. 

\begin{table}
\caption{Comparison of predicted and ground truth ODEs from ODEBench.}
\label{tab:case_study_odebench}
\centering
\resizebox{\columnwidth}{!}{%
\begin{tabular}{p{4cm}|p{5cm}|p{5cm}}
    \toprule
    ODE Description & Ground truth & Predicted \\
    \midrule
    Autocatalysis with one fixed abundant chemical  & 
    \[
    \frac{dx_0}{dt} = 2.1 \times x_0 - 0.5 \times x_0^2
    \] & 
    \[
    \frac{dx_0}{dt} = 2.113 \times x_0 - 0.504 \times x_0^2
    \] \\
    \midrule
    Harmonic oscillator with damping & 
    \[
    \begin{aligned}
        \frac{dx_0}{dt} &= x_1 \\
        \frac{dx_1}{dt} &= -4.5 \times x_0 - 0.43 \times x_1
    \end{aligned}
    \] & 
    \[
    \begin{aligned}
        \frac{dx_0}{dt} &= 1.099 \times x_1 \\
        \frac{dx_1}{dt} &= -4.699 \times x_0 - 0.466 \times x_1
    \end{aligned}
    \] \\
    \midrule
    SIR infection model only for healthy and sick &
    \[
    \begin{aligned}
        \frac{dx_0}{dt} &= -0.4 \times x_0 \times x_1 \\
        \frac{dx_1}{dt} &= -0.314 \times x_1 + 0.4 \times x_0 \times x_1
    \end{aligned}
    \] & 
    \[
    \begin{aligned}
        \frac{dx_0}{dt} &= -0.39 \times x_0 \times x_1 \\
        \frac{dx_1}{dt} &= -0.34 \times x_1 + 0.408 \times x_0 \times x_1
    \end{aligned}
    \] \\
    \midrule
    Lorenz equations in well-behaved periodic regime &
    \[
    \begin{aligned}
        \frac{dx_0}{dt} &= 5.1 \times x_1 - 5.1 \times x_0 \\
        \frac{dx_1}{dt} &= - x_1 + 12 \times x_0 - x_0 \times x_2 \\
        \frac{dx_2}{dt} &= - 1.67 \times x_2 + x_0 \times x_1
    \end{aligned}
    \] & 
    \[
    \begin{aligned}
        \frac{dx_0}{dt} &= 3.442 \times x_1 - 3.678 \times x_0 \\
        \frac{dx_1}{dt} &= 1.84 \times x_2 - 0.311 \times x_0^3 \\
        \frac{dx_2}{dt} &= - 4.043 \times x_2 + 10.738 \times x_0 
    \end{aligned}
    \] \\
    \bottomrule
    \end{tabular}%
}
\end{table}

\begin{table}
    \scriptsize
    \caption{Anesthesia surgery variables, their descriptions, and predicted ODEs.}
    \begin{tabularx}{\textwidth}{ c p{1.5cm} p{3.15cm} X }
        \toprule
        & \textbf{Name} & \textbf{Description} & \textbf{Predicted ODE} \\
        \midrule
        \textbf{Prop} & Propofol & An induction drug used & 
            \( \dot{\text{Prop}} = -0.0019 \times \text{Prop}^2 \) \\
        \midrule
        \textbf{HR} & Heart rate & Number of heart beats & 
            \( \dot{\text{hr}} = 0.0021 \times (1 - 0.0131 \times \text{hr})^2 \times (0.2098 - \text{Prop})^2 \) \\
        \midrule
        \textbf{sbp} & Systolic blood pressure & The pressure in the arteries when the heart contracts and pumps blood out & 
            \( \dot{\text{sbp}} = 0.3635 \times \text{mbp} - 0.2282 \times \text{sbp} \) \\
        \midrule
        \textbf{mbp} & Mean blood pressure & The average pressure in the arteries during a single cardiac cycle & 
            \( \dot{\text{mbp}} = 0.0618 \times \text{hr} - 0.1132 \times \text{BIS} \) \\
        \midrule
        \textbf{dbp} & Diastolic blood pressure & The pressure in the arteries when the heart is at rest between beats & 
            \( \dot{\text{dbp}} = 0.0382 \times \text{sbp} - 0.0824 \times \text{dbp} \) \\
        \midrule
        \textbf{BIS} & Bispectral Index & An index used to measure the depth of anesthesia & 
            \( \dot{\text{BIS}} = 0.0003 \times \text{mbp}^2 - 0.0603 \times \text{BIS} \) \\
        \bottomrule
    \end{tabularx}
    \label{tab:anes_vars}
\end{table}

%% file: section/5-Conclusion.tex
\section{Conclusion}
In this work, we introduce the divergence difference (DIV-diff) metric to improve the evaluation of predicted ODEs. We propose DDOT, a derivative-directed dual-decoder transformer model that predicts derivatives at each trajectory point to better approximate the true ODE vector field, resulting in more accurate divergence estimates. Experiments show DDOT outperforms the state-of-the-art ODEFormer in reconstruction, generalization, and divergence evaluation. 
We also demonstrate DDOT's application on an anesthesia surgery dataset, showing its potential for clinical decision support through interpretable symbolic representations. DIV-diff is a robust evaluation tool that can advance ODE system research by addressing initial value sensitivity and assessing performance across the variable space. 

%% file: main.bbl
\begin{thebibliography}{10}
\providecommand{\url}[1]{\texttt{#1}}
\providecommand{\urlprefix}{URL }
\providecommand{\doi}[1]{https://doi.org/#1}

\bibitem{DBLP:conf/ppsn/AgapitosBO12}
Agapitos, A., Brabazon, A., O'Neill, M.: Controlling overfitting in symbolic regression based on a bias/variance error decomposition. In: {PPSN} {(1)}. Lecture Notes in Computer Science, vol.~7491, pp. 438--447. Springer (2012)

\bibitem{BackToTheFormula}
Butter, A., Plehn, T., Soybelman, N., Brehmer, J.: Back to the formula-lhc edition. SciPost Physics  \textbf{16}(1), ~037 (2024)

\bibitem{LinearAlgebraWithTransformers}
Charton, F.: Linear algebra with transformers. Trans. Mach. Learn. Res.  \textbf{2022} (2022)

\bibitem{NODE}
Chen, T.Q., Rubanova, Y., Bettencourt, J., Duvenaud, D.: Neural ordinary differential equations. In: NeurIPS. pp. 6572--6583 (2018)

\bibitem{ODEFormer}
d'Ascoli, S., Becker, S., Schwaller, P., Mathis, A., Kilbertus, N.: Odeformer: Symbolic regression of dynamical systems with transformers. In: {ICLR}. OpenReview.net (2024)

\bibitem{DeepAR}
Flunkert, V., Salinas, D., Gasthaus, J.: Deepar: Probabilistic forecasting with autoregressive recurrent networks. CoRR  \textbf{abs/1704.04110} (2017)

\bibitem{E2ESR}
Kamienny, P., d'Ascoli, S., Lample, G., Charton, F.: End-to-end symbolic regression with transformers. In: NeurIPS (2022)

\bibitem{koza1994genetic}
Koza, J.R.: Genetic programming as a means for programming computers by natural selection. Statistics and computing  \textbf{4},  87--112 (1994)

\bibitem{DLforSymbolicMath}
Lample, G., Charton, F.: Deep learning for symbolic mathematics. In: {ICLR}. OpenReview.net (2020)

\bibitem{PMLB}
Olson, R.S., La~Cava, W., Orzechowski, P., Urbanowicz, R.J., Moore, J.H.: Pmlb: a large benchmark suite for machine learning evaluation and comparison. BioData Mining  \textbf{10}(1), ~36 (Dec 2017)

\bibitem{DeepSymbolicRegression}
Petersen, B.K., Landajuela, M., Mundhenk, T.N., Santiago, C.P., Kim, S., Kim, J.T.: Deep symbolic regression: Recovering mathematical expressions from data via risk-seeking policy gradients. In: {ICLR}. OpenReview.net (2021)

\bibitem{rudin2019stop}
Rudin, C.: Stop explaining black box machine learning models for high stakes decisions and use interpretable models instead. Nature Machine Intelligence  \textbf{1}(5),  206--215 (2019)

\bibitem{schmidt2009distilling}
Schmidt, M., Lipson, H.: Distilling free-form natural laws from experimental data. Science  \textbf{324}(5923),  81--85 (2009)

\bibitem{AI-feynman}
Udrescu, S., Tan, A.K., Feng, J., Neto, O., Wu, T., Tegmark, M.: {AI} feynman 2.0: Pareto-optimal symbolic regression exploiting graph modularity. In: NeurIPS (2020)

\bibitem{AttentionIsAllYouNeed}
Vaswani, A., Shazeer, N., Parmar, N., Uszkoreit, J., Jones, L., Gomez, A.N., Kaiser, L., Polosukhin, I.: Attention is all you need. In: {NIPS}. pp. 5998--6008 (2017)

\bibitem{SciPy}
Virtanen, P., Gommers, R., Oliphant, T.E., Haberland, M., Reddy, T., Cournapeau, D., Burovski, E., Peterson, P., Weckesser, W., Bright, J., {van der Walt}, S.J., Brett, M., Wilson, J., Millman, K.J., Mayorov, N., Nelson, A.R.J., Jones, E., Kern, R., Larson, E., Carey, C.J., Polat, {\.I}., Feng, Y., Moore, E.W., {VanderPlas}, J., Laxalde, D., Perktold, J., Cimrman, R., Henriksen, I., Quintero, E.A., Harris, C.R., Archibald, A.M., Ribeiro, A.H., Pedregosa, F., {van Mulbregt}, P., {SciPy 1.0 Contributors}: {{SciPy} 1.0: Fundamental Algorithms for Scientific Computing in Python}. Nature Methods  \textbf{17},  261--272 (2020)

\end{thebibliography}
